\documentclass{article}

\usepackage{microtype}
\usepackage{graphicx}
\usepackage{amsthm}
\usepackage{amssymb}
\usepackage{mathtools}
\usepackage{booktabs} 
\usepackage{multirow}
\usepackage{amsmath}
\usepackage{subcaption}
\usepackage[table]{xcolor}

\usepackage{hyperref}

\newenvironment{noindlist}
 {\begin{list}{\labelitemi}{\leftmargin=0em \itemindent=1em}}
 {\end{list}}
\newcommand{\vect} [1] {\boldsymbol{#1}}

\makeatletter
\newcommand{\subalign}[1]{%
	\vcenter{%
		\Let@ \restore@math@cr \default@tag
		\baselineskip\fontdimen10 \scriptfont\tw@
		\advance\baselineskip\fontdimen12 \scriptfont\tw@
		\lineskip\thr@@\fontdimen8 \scriptfont\thr@@
		\lineskiplimit\lineskip
		\ialign{\hfil$\m@th\scriptstyle##$&$\m@th\scriptstyle{}##$\hfil\crcr
			#1\crcr
		}%
	}%
}
\makeatother

\theoremstyle{remark}

\theoremstyle{theorem}
\newtheorem{theorem}{Theorem}

\theoremstyle{definition}
\newtheorem{definition}{Definition}

\usepackage[accepted]{icml2020}

\icmltitlerunning{Isometric Graph Neural Networks}

\begin{document}

\renewcommand{\algorithmicinput}{\textbf{Input:}}
\renewcommand{\algorithmicoutput}{\textbf{Output:}}

\twocolumn[
\icmltitle{Isometric Graph Neural Networks}



\icmlsetsymbol{equal}{*}

\begin{icmlauthorlist}
	\icmlauthor{Matthew Walker}{}
	\icmlauthor{Bo Yan}{}
	\icmlauthor{Yiou Xiao}{}
	\icmlauthor{Yafei Wang}{}
	\icmlauthor{Ayan Acharya }{}
	\\ \textbf{LinkedIn}
\end{icmlauthorlist}



\icmlkeywords{Machine Learning, Graph Neural Network}
\vskip 0.3in
]




\vspace{-0.5cm}
\begin{abstract}
Many tasks that rely on representations of nodes in graphs would benefit if those representations were faithful to distances between nodes in the graph. Geometric techniques to extract such representations have poor scaling over large graph size, and recent advances in Graph Neural Network (GNN) algorithms have limited ability to reflect graph distance information beyond the first degree neighborhood. To enable this highly desired capability, we propose a technique to learn \textbf{Isometric Graph Neural Networks} (IGNN), which requires changing the input representation space and loss function to enable \emph{any} GNN algorithm to generate representations that reflect distances between nodes. We experiment with the isometric technique on several GNN architectures for modeling multiple prediction tasks on multiple datasets. In addition to an improvement in AUC-ROC as high as $43\%$ in these experiments, we observe a consistent and substantial improvement as high as $400\%$ in Kendall's Tau (KT), a measure that directly reflects distance information, demonstrating that the learned embeddings do account for graph distances.

\end{abstract}

\vspace{-0.5cm}
\section{Introduction}

Position information plays an important role in many representation learning architectures. In modeling sequence data, such as sentences~\cite{sutskever2014sequence} and trajectories~\cite{alahi2016social}, the position information of each word and stay point is explicitly captured in the model design with recurrent patterns and mechanisms that preserve long-range sequential structure. Such modeling strategies are based upon the intuitive understanding of the temporal translation invariance and relational inductive bias~\cite{battaglia2018relational} commonly observed in sequence data. Studies~\cite{yang2016position,zeng-etal-2014-relation} have also shown that architectures, such as the Convolutional Neural Networks (CNN), specifically designed for 2D data, such as images, are able to outperform sequence models on natural language processing tasks when position features are encoded. Moreover, by modifying the CNN structure to specifically propagate position (spatial) information, performance is significantly improved on particular object detection tasks (e.g., lane detection) that require a better understanding of the spatial relationship of different pixels~\cite{pan2018spatial}.

This importance applies to learning representations in graphs which has gained a lot of traction among researchers and practitioners alike for the abundance and diversity of graph-structured data, such as molecule networks, social networks, and knowledge graphs. These representation learning approaches can be usually classified into three categories -- factorization-based approaches~\cite{belkin2002laplacian, ahmed2013distributed, ou2016asymmetric}, random walk-based approaches~\cite{grover2016node2vec, perozzi2014deepwalk}, and neighborhood aggregation and convolution-based approaches~\cite{Kipf:2016tc, hamilton2017inductive}. 

Among them, the Graph Neural Networks (GNN) algorithms that perform neighborhood aggregation and convolution have enjoyed a lot of popularity due to their superior performance on node classification, link prediction, and graph generation tasks~\cite{Kipf:2016tc, hamilton2017representation, liao2019efficient}. The key idea of GNN is to recursively gather information from neighboring nodes (a.k.a. message passing~\cite{gilmer2017neural}), in order to capture multi-hop dependencies by means of aggregation functions and non-linear transformations.

Due to the permutation invariance and isomorphism characteristics in graphs, certain aggregation operations, such as summation, are proven, both theoretically~\cite{xu2018how} and practically~\cite{hamilton2017inductive, hamilton2017representation}, to work better than the others. However, such findings do not preclude the possibility of improving the performance further by incorporating the position or relative ordering\footnote{We use concepts such as (relative) position, ordering, and distance interchangeably in the paper.} of nodes in the graph. On the other hand, \citet{xu2018how, DBLP:conf/icml/YouYL19} observe that GNNs are unable to learn distinct representations of the nodes that reside in different parts of the graph but have identical topological neighborhoods. Therefore, we seek to learn embeddings that are representative of relative distances on the graph, that is, are nearly isometric.

Numerous practical applications of graph embedding algorithms would also benefit if the learned embeddings respect the relative positions of the nodes in the original graph. One obvious example is that when performing nearest neighbor search over embeddings learned by GNNs, we might like the results to be actual neighbors in the graph sense. In an ontology, preserving the partial order of nodes in the embedding space is usually helpful for reasoning in the latent space~\cite{mcfee2009partial, DBLP:journals/corr/VendrovKFU15}. Many use cases of graph data, such as predicting career trajectories in a professional social network~\cite{yan2019time}, may also require the preservation of the explicit ordering of nodes, e.g., based on timestamps. Moreover, \citet{velivckovic2020neural} show that GNNs can execute many classical graph algorithms, such as breadth-first search, depth-first search, and shortest path algorithms, based on shared subroutines. The relative position of nodes and graph distances are important concepts within those subroutines, thus discovering embedding algorithms that may preserve graph distances can facilitate further research in this area. Inspired by these findings, we investigate the hitherto unexplored connection between relative node positions/graph distances and the expressive power of GNNs. 

Though the positions of the nodes can be explicitly captured in Position-aware GNN (P-GNN) models with the assistance of anchor nodes sampled randomly from the graph~\cite{DBLP:conf/icml/YouYL19}, such an approach requires a lot of preprocessing (e.g., anchor set selection), introduces additional communications between the nodes and the anchor sets, and is not practically feasible for graphs that evolve rather frequently. On the other hand, simple module-like encoding mechanisms of position information work extremely well for sequence data~\cite{vaswani2017attention, xiao2014fast} and spatial feature representation learning~\cite{mai2020multiscale}. Motivated by these works, we propose to incorporate a hashing mechanism to encode the position information of the nodes in the graph. Such a hashing mechanism is independent of any specific GNN algorithm and hence can enhance \emph{any} GNN architecture. We justify the utility of the hashing module for encoding positions in the graph by providing a theoretical connection between hashing functions and isometric (distance-preserving) embeddings.  Since neural networks can implicitly learn spatial positioning in images~\cite{islam2020how} and navigation tasks~\cite{j.2018emergence, banino2018vector, gao2018learning}, we further investigate whether the GNNs are capable of utilizing the position signals if such a preference is encoded explicitly in the learning objective. The introduction of these two extensions -- hashed feature augmentation and the inclusion of the position information in the learning objective -- begets \textbf{Isometric Graph Neural Networks (IGNN)}. Note that P-GNN tries to minimize distortion and break local isomorphism in a probabilistic manner. In contrast, IGNN addresses these issues using a deterministic formulation that can work with any GNN architecture (see Section~\ref{sec:isometric}).  

To summarize, the main contributions of our work are the following:
\vspace{-0.3cm}
\begin{noindlist}
	\item We connect and expand theoretical results on finding isometric embeddings to graph neural network techniques.
	\vspace{-0.2cm}
	\item We provide a prescription for learning near-isometric embeddings using any graph neural network algorithm.
	\vspace{-0.2cm}
	\item We demonstrate that applying this technique substantially improves performance in metrics sensitive to network distance in a number of datasets and prediction tasks.
\end{noindlist}
\vspace{-0.5cm}
\section{Related Work}
Existing research on GNN has primarily focused on the expressiveness and representational power of different embedding methods. Since the underlying mechanism of GNN variants relies on the message passing and aggregation procedures~\cite{gilmer2017neural}, a large amount of effort has been invested to discover different aggregation functions to achieve state-of-the-art performance on node classification, link prediction, and graph generation tasks. Empirically, permutation invariant aggregation functions (e.g., mean, sum, max-pooling, min-pooling) are ideal choices for GNNs to learn embeddings in general~\cite{hamilton2017representation, hamilton2017inductive, Kipf:2016tc}. Theoretically, \citet{xu2018how} have established the intricate connection between the discriminative power of GNN and the powerful Weisfeiler-Lehman (WL) graph isomorphism test~\cite{weisfeiler1968reduction} used to distinguish different graph structures while \citet{chen2019equivalence} have shown an explicit connection between the graph isomorphism test and permutation invariant function approximation.

Besides graph structures, the position, the ordering, and the relative graph distance of nodes have also been taken into consideration for designing expressive GNN architectures as such information is critical in distinguishing topologically identical substructures. Graph kernels~\cite{yanardag2015deep, sugiyama2015halting} have been used to encode position information in graph representation learning. \citet{DBLP:conf/icml/YouYL19} proposed the P-GNN model that incorporates additional message communication using anchor nodes. Compared with graph kernel methods, P-GNN models are closely related to and can be viewed as a generalization of existing GNN models. On a related note, \citet{maron2019provably} explored and developed GNN models that incorporated standard Multi-Layer Perceptrons (MLPs) of the feature dimension and a matrix multiplication layer and proved that they have higher expressiveness compared with message passing GNNs while maintaining scalability.

While the goal of these prior works is to develop more expressive GNN models using position information, little theoretical understanding of the mechanism in preserving graph distance (relative node positions) using GNN models has been established. Here we aim to provide theoretical connections among graph distance preservation, hashing functions, and the learning power of GNNs. Besides, the proposed method can be easily integrated with different GNN architectures to suit different needs.
\section{Preliminaries}
Let $\mathcal{G} = (\mathcal{V}, \mathcal{E})$ be a graph with a set of vertices $\mathcal{V}$ and a set of edges $\mathcal{E}$, such that $\forall (v_i, v_j) \in \mathcal{E}| v_i, v_j \in \mathcal{V}$. Let $N = |\mathcal{V}|$. When we refer to the graph distance $d_{\mathcal{G}}(v_i,v_j)$, we mean the shortest path on graph edges between nodes $v_i$ and $v_j$.

\textbf{Graph Neural Networks.}\label{preliminiaries} Graph Neural Network algorithms learn embeddings for the nodes in a graph using both graph structure and node specific features $\vect{X} = (\mathbf{x}_{v_i})_{v_i}$, where $\mathbf{x}_{v_i} \in \mathcal{X}$ is the node feature vector corresponding to node $v_i$. These algorithms learn a function, $f \vcentcolon \mathcal{V} \times \mathcal{X} \rightarrow \mathcal{Z}$, allowing nodes in $\mathcal{V}$ to be mapped to real-valued vectors $\mathbf{z} \in \mathcal{Z}$.

GNN algorithms define a neighborhood for each node in the graph and aggregates messages from these neighbors to update the representation of the current node. Formally, these algorithms iteratively update the representation of a node $v_i$ by performing aggregation and combinations on the neighborhood $\mathcal{N}(v_i)$ (the definition of which varies among the algorithms) for $K$ iterations. The aggregation function $\mathbf{g}^{k}_{v_i} = \textsc{agg}^k (\{\mathbf{z}^{(k-1)}_{v_j} | v_j \in \mathcal{N}(v_i)\})$ performs the aggregation over a set of vectors $\{\mathbf{z}^{(k-1)}_{v_i}\}$ for each neighbor of node $v_i$ from iteration $(k-1) < K$ to obtain the neighborhood representation of node $v_i$ for iteration $k \leq K$. Typical permutation invariant choices for $\textsc{agg}$ are mean, max-pooling, and sum~\cite{Kipf:2016tc, hamilton2017inductive, xu2018how}. The combination function $\textsc{comb}^k(\mathbf{z}^{(k-1)}_{v_i}, \mathbf{g}^{k}_{v_i})$ combines the node representation $\mathbf{z}^{(k-1)}_{v_i}$ for $v_i$ at iteration $(k-1)$ with the neighborhood representation $\mathbf{g}^{k}_{v_i}$ for $v_i$ at iteration $k$. Typical choices for $\textsc{comb}$ are concatenation and summation. In general, a GNN can be represented as 
\begin{equation}\label{eq:gnnConv}
\resizebox{.8\hsize}{!} {$\mathbf{z}^k_{v_i} = f\big((\textsc{comb}^k \circ \textsc{agg}^k) (\mathbf{z}^{(k-1)}_{v_i}, \{\mathbf{z}^{(k-1)}_{v_j} | v_j \in \mathcal{N}(v_i)\})\big)$}
\end{equation}
where function $f$, here, is usually a feed forward neural network (such as a single-layer NN or an MLP). The order in which $\textsc{agg}$ and $\textsc{comb}$ are applied may be reversed, such as in the case of Graph Convolutional Networks (GCN)~\cite{Kipf:2016tc}. The output of a GNN is the $K$-th representation of each node, that is, $\mathbf{z}_{v_i} \vcentcolon= \mathbf{z}^K_{v_i} , \forall v_i \in \mathcal{V}$. Therefore, in GNN algorithms, the parameter set is defined by the formulation of the aggregation, combination, and $f$ functions.


\textbf{Representation Spaces.} Since our objective is to learn representations that are reflective of structural similarity, graph distance, and the similarity inferred by the node features, it is important to outline the differences among various types of spaces we incorporate distance measures from. How they relate, practically, to implementations of GNNs vary, as certain theoretical results apply to different types of spaces.

\begin{definition}
A \textit{metric space} $(\mathcal{X}, d)$ is a set $\mathcal{X}$ endowed with a \textit{distance function} $d: \mathcal{X}\times \mathcal{X}\rightarrow \mathbb{R}$, sometimes called a \textit{metric}. This distance function  satisfies three key properties for any $x, y, z \in \mathcal{X}$:  positivity ($d(x,y) \geq 0$), symmetry ($d(x,y) = d(y,x)$), and triangle inequality ($d(x,z) \leq d(x,y) + d(y,z)$). \end{definition}

\begin{definition}
A \textit{finite} metric space is a metric space with a finite number of points.
\end{definition}

\begin{definition}
 A \textit{Euclidean space} is a metric space with $\mathcal{X}=\mathbb{R}^n, n\in \mathbb{Z}_{+}$ characterized by a distance function $d(\mathbf{x}, \mathbf{y}) = (\sum^n_{i=1} (x_i - y_i)^2)^{1/2}, \forall \mathbf{x}, \mathbf{y} \in \mathbb{R}^n$.
\end{definition}

\begin{definition}
\label{definition:embedding}
An \textit{embedding} is defined as a map $f \vcentcolon \mathcal{X} \rightarrow \mathcal{X}^\prime$ between two metric spaces $(\mathcal{X}, d)$ and $(\mathcal{X}^\prime, d^\prime)$.
\end{definition}

\begin{definition}
\label{definition:isometric}
An embedding is \textit{isometric} iff $\forall x, y \in \mathcal{X}, d(x,y) = d^\prime(f(x), f(y))$.
\end{definition}
Since graphs with weighted edges are not guaranteed to satisfy the triangle inequality, they are generally not isomorphic to finite metric spaces or the Euclidean space.

\begin{definition}
\label{definition:distortion}
The \emph{distortion} of an embedding is the smallest value $\alpha \geq 1$ for which there exists an $r > 0$ such that $\forall x, y \in \mathcal{X}$,
\begin{equation}\label{eq:distortion}
r~d(x,y) \leq d^\prime(f(x), f(y)) \leq \alpha~r~d(x,y)
\end{equation}
\end{definition}


\textbf{Finding Isometric Embeddings.} The topic of finding isometric and near-isometric embeddings between spaces is a well-studied problem. The study of embeddings of finite metric spaces into Euclidean spaces is of particular interest to the problem of finding representations for nodes in the graph. To that end, Theorem~\ref{theorem:bourgain}~\cite{Bourgain1985,linalLondonRabinovichGraphGeometry} has been studied in relation to graph neural networks~\cite{DBLP:conf/icml/YouYL19, srinivasan2019equivalence}. Furthermore, the proof of Theorem~\ref{theorem:bourgain} is algorithmic, guiding how such an embedding might be constructed.

\begin{theorem}
\label{theorem:bourgain}
(Bourgain Theorem) Every n-point metric space ($\mathcal{X}$, d) can be embedded in an $\mathcal{O}$(\textrm{log} $n$)-dimensional Euclidean space with $\mathcal{O}$(log $n$) distortion.
\end{theorem}

This promising result is extended by~\citet{linalLondonRabinovichGraphGeometry} who show that such an embedding produces the least possible distortion. Unfortunately, the algorithm to find such an embedding takes random polynomial time. Moreover, only when the embedding is achieved on an $\mathcal{O}$($n^2$)-dimensional space one can adopt a deterministic polynomial-time algorithm. Therefore, both of these alternatives are unappealing for learning representations of large graphs. P-GNN~\cite{DBLP:conf/icml/YouYL19} takes inspiration from this approach, but as the distortion is probabilistic, a given choice of anchor nodes provides no guarantee on the level of distortion. Furthermore, an embedding learned on the input feature space has a distance measure with no connection, in general, to the graph distance.

\section{Isometric Graph Neural Network}\label{sec:isometric}

This section presents the mathematical details of the \textbf{IGNN} framework. As explained in the previous section, in general, it is not possible to learn a representation that preserves the exact distance in a graph. Therefore, for any practical application, one must find an embedding that minimizes the distortion. To that end, we first propose a modular extension of the input representation of the graph data using hashed features. In what follows, we concretely argue for the utility, scalability and injectiveness of such modification of the input features. To reduce the distortion further, we propose a modification to the loss function that explicitly penalizes for large deviations in the embedding space from the distance dictated by the graph. Finally, we discuss the complexity of adopting this prescription during model training.

\textbf{Input Representation.} Existing GNN algorithms encode the features of an individual node in two different forms -- as a set of dataset-dependent attributes or as a one-hot encoded vector that acts as a proxy for the node identity. Such representations can be realized in a Euclidean metric space trivially, though the distance function associated with such metric space, in general, is completely uncorrelated with the distance inferred by the graph. Therefore, learning an embedding from such representation of the features may not preserve the distances in the graph. Given that any linear transformation $f(\mathbf{x}) = \mathbf{A}^{T}\mathbf{x}$, where $\mathbf{A}$ is an orthogonal matrix, is an isometric embedding, it is extremely easy to find isometric embeddings between Euclidean spaces, but difficult to find an embedding that respects the distances in the original graph. 

To that end, we propose to extend the input feature representations of the nodes in the GNNs by concatenating them with a hash vector. The objective of this enhancement is to increase the dimensionality of the input space so that it helps the GNNs break isomorphisms in the graph and learn a representation that is sensitive to the graph distance. In what follows, we first describe the construction of the hash vector and then argue about the injectiveness it introduces that helps us achieve the objectives mentioned above. 

\begin{definition}
\label{definition:hashvector}
A \emph{hash vector ($n$, $m$)} is an $n$-dimensional vector, where each dimension is an $m$-bit string generated by a hash function and input seed.
\end{definition}

In Algorithm~\ref{algo:hashvec}, we describe how a hash vector can be generated using a hash function that produces $m$-bit strings from an input bit string. We show in Theorem~\ref{theorem:hash} that this algorithm provides protection from collision as good as generating the hash from $n$ independent seeds.

\begin{algorithm}
\begin{algorithmic}
\STATE \algorithmicinput~A hash function $\mathcal{H}_m\vcentcolon \{0,1\}^* \rightarrow \{0,1\}^m$; an integer $n$ defining the dimension of the output vector; the input bit string $b$ which generates the vector.
\STATE \algorithmicoutput~A hash vector $h \in \{0,1\}^{m\times n}$
\STATE $h \leftarrow (~)$
\FOR{$i=0$ until $n$}
\STATE $t \leftarrow \mathcal{H}(b \oplus i) $; $h \leftarrow h \oplus t$;
\ENDFOR
\end{algorithmic}
\caption{Constructing a hash vector}
\label{algo:hashvec}
\end{algorithm}

\begin{theorem}
\label{theorem:hash}
Given a hash function $\mathcal{H}_m: \{0,1\}^* \rightarrow \{0,1\}^m$ with a collision probability $p \vcentcolon= p(\mathcal{H}(x) = \mathcal{H}(y) | x \neq y)$, the collision probability between two hash vectors of dimension $n$, generated by Algorithm~\ref{algo:hashvec}, is $p^n$.
\end{theorem}

\emph{Proof:} Consider two hash vectors $\{h_{i,1}\}$ and $\{h_{i,2}\}$, $0 \leq i \leq n$, generated by input bit strings $b_1 \neq b_2$. These hash vectors collide if $h_{i,1} = h_{i,2}, \forall i$. Given that $h_{i} = \mathcal{H}(h_{i-1} \oplus b)$, the vector components $h_i$ are randomly distributed in the hash space. Therefore, the probability of collision is independent for each dimension in the hash vector.

Note that adding the hash vector representation introduces a high degree of injectiveness into the input representation. \citet{xu2018how,maron2019provably} discuss at length how introducing some form of injectiveness into GNN algorithms allows one to break the local structural isomorphisms that prevail in the underlying graph. Compared to these existing approaches that either have probabilistic guarantees or no guarantees on injectiveness, augmentation of the hash representation offers a guarantee of injectiveness up to the collision probability of the hash function and hash vector size. This improvement applies both to datasets with node features and those using one-hot encoded vectors, for which it is very easy to generate unique input seeds for each node.

The approach has several additional advantages. Since the hashes have no security constraints, one may use extremely fast collision-resistant hash functions available for a scalable implementation. The process is also deterministic, guaranteeing that the input representation is reproducible across training runs, inference, and other analysis tasks.

%
%
%
%

\textbf{Learning the Embedding.} The Johnson-Lindenstrauss Lemma~\cite{johnson1984lipschitz}, is a weaker result than Theorem~\ref{theorem:bourgain} (Bourgain Theorem). As described by~\citet{linalLondonRabinovichGraphGeometry}, the proof is prescriptive, but unfortunately, is still a random polynomial time algorithm. An alternative proof is provided by~\citet{frankl1988}. The proof shows that there is a random subspace of the original space that satisfies the distortion requirement because the length of projections on random subspaces is normally distributed.

\begin{theorem}
\label{theorem:distortion}
(Johnson-Lindenstrauss Lemma) 
Given $0< \epsilon < 1$, any set of n points in a Euclidean space can be mapped to $\mathbb{R}^q$ with a distortion $\alpha \leq (1 + \epsilon)$ in the distances if 
$q > 8 \log n /\epsilon^2$.
\end{theorem}

Though the algorithm for which the lemma holds does not scale, it strongly suggests that one can find a mapping between two Euclidean spaces with minimal distortion, which is what is relevant to most GNN architectures. Because the hash vectors add random projections into a new space, the appropriate embedding learned on this space can minimize the distortion with respect to the graph distance. Recent exceptions~\cite{ChamiHyperbolic2019,liuHyperbolic2019} focus on hyperbolic spaces rather than Euclidean, but appealing to the Nash Embedding Theorem~\cite{nashEmbeddingTheorem} extends the arguments of this paper to those architectures.

In order to ensure that the mapping learned by the GNN algorithm is minimally distorted with respect to the graph distance metric, the graph distance must be explicitly incorporated into the training objective. In Section \ref{exp}, we elaborate on how the loss function must be modified to achieve such an objective.

\textbf{Complexity Analysis.} Since IGNN augments the existing GNN architectures with hashed features, it behooves us to examine the complexity difference compared to these architectures. There are three sources of complexity changes when adopting this prescription. First, calculating the hash vectors incurs a cost of $\mathcal{O}(Nn)$, where $N$ is the number of nodes and $n$ is the dimension of the hash vectors. Adding the hash vectors introduces $\mathcal{O}(n)$ additional model parameters, which introduce additional cost during both inference and back-propagation. There is a fixed cost of $\mathcal{O}(N^3)$ of calculating the graph distances between the nodes for use in the loss function.\footnote{The same cost is needed in P-GNN models.} For a large graph, one may use an approximate shortest path algorithm \cite{hekn13} to avoid the cubic complexity.
\section{Experiment}
\label{exp}

\subsection{Methods}
\textbf{Extending Input Representation.} We extend the input representation (either one-hot vectors or node features) $\mathbf{x}_{v_i}$ for node $v_{i}$ with fixed-length hash vectors
\begin{equation}
\mathbf{x}^\mathcal{H}_{v_i}=[\mathcal{H}(b_i\oplus 0), \mathcal{H}(b_i\oplus 1),\dots,\mathcal{H}(b_i\oplus n-1)]^T
\end{equation}
where $b_i$ is an unique identifier for each node to achieve determinism and $n$ denotes the desired length of the hash features. In all of our experiments, we adopt
\begin{equation}
\mathcal{H}(\cdot)=\text{MurmurHash3}(\cdot)/(2^{31}-1)
\end{equation}
as the default hashing algorithm \cite{applebymurmurhash3} which hashes an arbitrary string into the range $[-2^{31}-1, 2^{31}-1]$. In essence, we provide a modular extension to all GNN algorithms by extending the input representation of each node to $\mathbf{x}_{v_i}^{*} = \text{CONCAT}(\mathbf{x}_{v_i}, \mathbf{x}^\mathcal{H}_{v_i})$. In contrast to the original input representation of the nodes $\mathbf{x}_i$ that encode application-specific characteristics, the features $\mathbf{x}_{v_i}^\mathcal{H}$ derived from hashing, spanning the space $[-1, 1]^n$, are designed to counter-balance the isomorphism between remote communities.

\textbf{Application Tasks.} We enable all GNN algorithms with the extended input representations to learn a mapping $f:\mathcal{V}\times \mathcal{X}\rightarrow\mathcal{Z}$ based on the target similarity metric $d_y(\mathbf{z}_{v_i,}\mathbf{z}_{v_j})$ determined by downstream application tasks, such as link prediction, pairwise node classification and so on. In particular, we focus on the following two prediction tasks.
\begin{noindlist}
	\item Link Prediction: The objective here is to model a similarity metric defined by $d_y(\mathbf{z}_{v_i}, \mathbf{z}_{v_j}) = 1$ if $(v_i, v_j)\in\mathcal{E}^{+}$ and $0$ if $(v_i, v_j)\in\mathcal{E}^{-}$, where $\mathcal{E}^+\subset\mathcal{E}$ represents the set of positive edges and $\mathcal{E}^-=\mathcal{V}\times\mathcal{V} - \mathcal{E}$ represents the set of negative edges sampled from the underlying graph $\mathcal{G} = (\mathcal{V}, \mathcal{E})$. 
	\item Pairwise Node Classification: In this task, in addition to the graph $\mathcal{G}$, an exclusive node-set partition $\{\mathcal{P}_1, \mathcal{P}_2\dots,\mathcal{P}_M\}\subseteq \mathcal{V}$ is provided. The target similarity metric for this task is defined as $d_y(\mathbf{z}_{v_i}, \mathbf{z}_{v_j})= 1$ if $\exists \ell (v_i\in\mathcal{P}_\ell\wedge v_j\in\mathcal{P}_\ell)$ and $0$ otherwise.
\end{noindlist}
Note that the subscript $y$ in $d_y(.,.)$ makes it explicit that the similarity is calculated at the output task level. We will introduce two more distances of similar kind related to the graph  and the embedding space and hence it is important to make the notations for distances clear and distinct. 

\textbf{Optimization over Extended Representation.} To generalize the learning task introduced in Section~\ref{preliminiaries}, most GNN algorithms seek to learn a mapping $f(v_i, \mathcal{N}(v_i), \mathbf{x}_{v_i}, \{\mathbf{x}_{v_j}: v_j \in \mathcal{N}(v_i)\})$ where $\mathcal{N}(v_i)$ is taken as a collection of one-hop or two-hop neighbors for node $v_i$. P-GNN, on the other hand, constructs the neighborhood a little differently by augmenting the set $\mathcal{N}(v_i)$ with a set of anchor nodes $\mathcal{S}$. Note that this set of anchor nodes appear in the neighborhood of each node in the graph. In contrast, our approach seeks to learn a function with extended input representations that can be 
formally written as $f(v_i, \mathcal{N}(v_i), \mathbf{x}^{*}_{v_i}, \{\mathbf{x}^{*}_{v_j}: v_j \in \mathcal{N}(v_i)\})$.

As described by~\citet{DBLP:conf/icml/YouYL19}, most GNN models can be formalized by the following optimization:
\begin{equation}
\resizebox{0.9\hsize}{!}{$\min {\,\mathbb{E}_{v_i, v_j, \mathcal{N}(v_i), \mathcal{N}(v_j)}} [\mathcal{L}(d_z(f_{v_i}, f_{v_j}), d_y(\mathbf{z}_{v_i}, \mathbf{z}_{v_j}))]$}
\end{equation}
where $f_{v_i} = f(v_i, \mathcal{N}(v_i), \mathbf{x}_{v_i}, \{\mathbf{x}_{v_{\ell}}:v_{\ell} \in \mathcal{N}(v_i)\})$ and $f_{v_j}= f(v_j, \mathcal{N}(v_j), \mathbf{x}_{v_j}, \{\mathbf{x}_{v_{\ell}}: v_{\ell} \in \mathcal{N}(v_j)\})$. The random variables $v_i$, $v_j$, $\mathcal{N}(v_i)$, $\mathcal{N}(v_j)$ are all sampled from the probability distribution $p(\mathcal{V})$ over the set of vertices, dictated by some properties of the graph. Likewise, with a small modification to the neighborhood construction $\mathcal{N}(v_i)$ for each node (i.e. extending every neighborhood with the anchor nodes), PGNN can also be shown to optimize the objective given above. Note that the subscripts in $d_z(.,.)$ and $d_y(.,.)$ are meant to distinguish between the similarity metrics in the embedding space and target metric space respectively. 

In contrast, our approach optimizes for the composite loss function given by:
\begin{equation}
\resizebox{0.9\hsize}{!}{$\min {\,\mathbb{E}_{v_i, v_j, \mathcal{N}(v_i), \mathcal{N}(v_j)}} [\mathcal{L}(d_z(f^{*}_{v_i}, f^{*}_{v_j}), d_y(\mathbf{z}_{v_i}, \mathbf{z}_{v_j}), d_{\mathcal{G}}(v_i, v_j))]$}
\end{equation}
where $f^{*}_{v_i} = f(v_i, \mathcal{N}(v_i), \mathbf{x}^{*}_{v_i}, \{\mathbf{x}^{*}_{v_\ell}: v_\ell \in \mathcal{N}(v_i)\})$ and $f^{*}_{v_j}= f(v_j, \mathcal{N}(v_j), \mathbf{x}^{*}_{v_j}, \{\mathbf{x}^{*}_{v_\ell}: v_\ell \in \mathcal{N}(v_j)\})$. Note that we have augmented the loss function with an extra term $d_{\mathcal{G}}(v_i, v_j)$ that denotes the distance between node $v_i$ and $v_j$ according to the underlying graph $\mathcal{G}$. This explicit dependence on the graph distance is expected to guide the optimization to a point where the learned embeddings can have minimal distortion. Also note that the feature presentation in the above objective function accounts for the extended set of features.

To make the minimization problem above explicit, we propose the following objective function
\vspace{-0.1cm}
\begin{equation}\label{eq:objective}
\mathcal{L} = \lambda_{\textrm{BCE}}\mathcal{L}_{\textrm{BCE}} + \lambda_{\textrm{MSE}}\mathcal{L}_{\textrm{MSE}}
\vspace{-0.1cm}
\end{equation}
which has two parts -- the usual task-specific target similarity loss term, which is the regular binary cross entropy (BCE) loss, and the new distance metric loss term, which is a mean-squared error (MSE) term. The BCE loss can be formalized as: 
\vspace{-0.3cm}
\begin{multline}
\resizebox{0.84\hsize}{!}{$\mathcal{L}_{\text{BCE}} = - \displaystyle\sum_{v_i\neq v_j} [d_y(\mathbf{z}_{v_i}, \mathbf{z}_{v_j}) \log \sigma(\langle \mathbf{z}_{v_i}, \mathbf{z}_{v_j} \rangle)$} \\
\resizebox{0.84\hsize}{!}{$+ (1- d_y(\mathbf{z}_{v_i}, \mathbf{z}_{v_j})) \log (1-\sigma(\langle \mathbf{z}_{v_i}, \mathbf{z}_{v_j} \rangle))]$}
\vspace{-0.1cm}
\end{multline}
Similarly, the distance metric can be characterized as:	
\begin{equation}
\resizebox{0.9\hsize}{!}{$\mathcal{L}_{\text{MSE}} = \displaystyle\sum_{v_i\neq v_j} [(1 - \langle \mathbf{z}_{v_i}, \mathbf{z}_{v_j} \rangle)/2 - (1 - 1/d_{\mathcal{G}}(v_i, v_j)^\alpha)]^2 $}
\vspace{-0.1cm}
\end{equation}
where $\alpha$ is a hyper-parameter that controls the shape of the distance transformation. In all of our experiments, we set $\alpha = 1$. A larger value of $\lambda_{\textrm{MSE}}$ or $\lambda_{\textrm{BCE}}$ favors the distance metric loss or the binary cross-entropy loss respectively. Interestingly, depending on the actual training task (i.e., link prediction, pairwise node classification, etc), these two losses might positively correlate. For example, for link prediction tasks, there exists a clear positive correlation between $1-d_y(\mathbf{z}_{v_i}, \mathbf{z}_{v_j})$ and $(1 -1 /d_{\mathcal{G}}(v_i, v_j)^\alpha)$.


\begin{table*}[!htb]
	\caption{Results for Communities and Email Dataset}
	\label{Exp:all}
	\begin{center}
	\resizebox{0.97\linewidth}{!}{
	\begin{tabular}{ccccccccc}
		\toprule
		\multicolumn{1}{c}{\multirow{3}{*}{}} &
		\multicolumn{4}{c}{Communities} &
		\multicolumn{4}{c}{Email} \\ \cline{2-9} 
		\multicolumn{1}{c}{Model Variants} &
		\multicolumn{2}{c}{Link Prediction} &
		\multicolumn{2}{c}{Pairwise Node Classification} &
		\multicolumn{2}{c}{Link Prediction} &
		\multicolumn{2}{c}{Pairwise Node Classification} \\ \cline{2-9} 
		\multicolumn{1}{c}{} &
		\multicolumn{1}{c}{AUC} &
		\multicolumn{1}{c}{Kendall's Tau} &
		\multicolumn{1}{c}{AUC} &
		\multicolumn{1}{c}{Kendall's Tau} &
		\multicolumn{1}{c}{AUC} &
		\multicolumn{1}{c}{Kendall's Tau} &
		\multicolumn{1}{c}{AUC} &
		\multicolumn{1}{c}{Kendall's Tau} \\ \hline
		GCN &
		$0.977 \pm 0.005$ &
		$0.182 \pm 0.038$ &
		$0.988 \pm 0.001$ &
		$0.146 \pm 0.015$ &
		$0.709 \pm 0.010$ &
		$0.239 \pm 0.025$ &
		$0.518 \pm 0.010$ &
		$0.238 \pm 0.008$ \\
		GCN + Hash &
		$0.986 \pm 0.001$ &
		$0.215 \pm 0.001$ &
		$0.992 \pm 0.005$ &
		$0.217 \pm 0.002$ &
		$0.767 \pm 0.005$ &
		$0.364 \pm 0.008$ &
		$0.681 \pm 0.030$ &
		$0.432 \pm 0.002$ \\
		GCN + MSE &
		$0.986 \pm 0.002$ &
		$0.267 \pm 0.003$ &
		$0.992 \pm 0.002$ &
		$0.277 \pm 0.003$ &
		$0.721 \pm 0.005$ &
		$0.217 \pm 0.004$ &
		$0.575 \pm 0.020$ &
		$0.267 \pm 0.050$ \\
		GCN + Both &
		$0.986 \pm 0.003$ &
		\boldmath{$0.301 \pm 0.002$} &
		$0.992 \pm 0.002$ &
		\boldmath{$0.335 \pm 0.003$} &
		$0.782 \pm 0.016$ &
		\boldmath{$0.364 \pm 0.002$} &
		$0.708 \pm 0.020$ &
		\boldmath{$0.437 \pm 0.002$} \\
		\hline
		SAGE &
		$0.986 \pm 0.003$ &
		$0.218 \pm 0.003$ &
		$0.993 \pm 0.001$ &
		$0.212 \pm 0.015$ &
		$0.571 \pm 0.033$ &
		$0.147 \pm 0.011$ &
		$0.538 \pm 0.007$ &
		$0.139 \pm 0.012$ \\
		SAGE + Hash &
		$0.979 \pm 0.004$ &
		\boldmath{$0.268 \pm 0.029$} &
		$0.982 \pm 0.006$ &
		$0.257 \pm 0.043$ &
		$0.735 \pm 0.000$ &
		$0.361 \pm 0.017$ &
		$0.693 \pm 0.010$ &
		$0.428 \pm 0.007$ \\
		SAGE + MSE &
		$0.985 \pm 0.003$ &
		$0.248 \pm 0.004$ &
		$0.994 \pm 0.002$ &
		$0.270 \pm 0.038$ &
		$0.557 \pm 0.069$ &
		$0.156 \pm 0.030$ &
		$0.533 \pm 0.001$ &
		$0.206 \pm 0.033$ \\
		SAGE + Both &
		$0.986 \pm 0.003$ &
		$0.253 \pm 0.002$ &
		$0.993 \pm 0.002$ &
		\boldmath{$0.280 \pm 0.002$} &
		$0.769 \pm 0.016$ &
		\boldmath{$0.363 \pm 0.002$} &
		$0.744 \pm 0.022$ &
		\boldmath{$0.433 \pm 0.002$} \\
				\hline
		GAT &
		$0.981 \pm 0.001$ &
		$0.203 \pm 0.028$ &
		$0.989 \pm 0.005$ &
		$0.205 \pm 0.006$ &
		$0.538 \pm 0.015$ &
		$0.086 \pm 0.009$ &
		$0.507 \pm 0.006$ &
		$0.083 \pm 0.018$ \\
		GAT + Hash &
		$0.919 \pm 0.008$ &
		$0.263 \pm 0.014$ &
		$0.975 \pm 0.004$ &
		\boldmath{$0.334 \pm 0.028$} &
		$0.758 \pm 0.010$ &
		$0.336 \pm 0.010$ &
		$0.725 \pm 0.017$ &
		$0.407 \pm 0.004$ \\
		GAT + MSE &
		$0.983 \pm 0.001$ &
		$0.300 \pm 0.069$ &
		$0.989 \pm 0.001$ &
		$0.318 \pm 0.004$ &
		$0.550 \pm 0.013$ &
		$0.096 \pm 0.001$ &
		$0.528 \pm 0.009$ &
		$0.112 \pm 0.001$ \\
		GAT + Both &
		$0.987 \pm 0.003$ &
		\boldmath{$0.303 \pm 0.002$} &
		$0.993 \pm 0.002$ &
		$0.324 \pm 0.002$ &
		$0.769 \pm 0.015$ &
		\boldmath{$0.353 \pm 0.002$} &
		$0.747 \pm 0.200$ &
		\boldmath{$0.435 \pm 0.002$} \\
		\hline
		GIN &
		$0.980 \pm 0.002$ &
		$0.241 \pm 0.010$ &
		$0.991 \pm 0.002$ &
		$0.213 \pm 0.041$ &
		$0.724 \pm 0.008$ &
		$0.402 \pm 0.004$ &
		$0.723 \pm 0.032$ &
		$0.479 \pm 0.001$ \\
		GIN + Hash &
		$0.988 \pm 0.007$ &
		$0.272 \pm 0.001$ &
		$0.982 \pm 0.006$ &
		$0.250 \pm 0.021$ &
		$0.785 \pm 0.007$ &
		$0.443 \pm 0.012$ &
		$0.741 \pm 0.001$ &
		\boldmath{$0.525 \pm 0.001$} \\
		GIN + MSE &
		$0.987 \pm 0.003$ &
		$0.290 \pm 0.002$ &
		$0.992 \pm 0.002$ &
		$0.301 \pm 0.002$ &
		$0.791 \pm 0.012$ &
		$0.429 \pm 0.001$ &
		$0.726 \pm 0.012$ &
		$0.482 \pm 0.001$ \\
		GIN + Both &
		$0.987 \pm 0.009$ &
		\boldmath{$0.432 \pm 0.092$} &
		$0.984 \pm 0.000$ &
		\boldmath{$0.365 \pm 0.070$} &
		$0.808 \pm 0.004$ &
		\boldmath{$0.452 \pm 0.005$} &
		$0.774 \pm 0.005$ &
		$0.521 \pm 0.007$ \\
				\hline
		P-GNN-F-2L &
		$0.978 \pm 0.007$ &
		$0.334 \pm 0.003$ &
		$0.986 \pm 0.001$ &
		$0.346 \pm 0.005$ &
		$0.751 \pm 0.040$ &
		$0.503 \pm 0.002$ &
		$0.751 \pm 0.013$ &
		$0.576 \pm 0.007$ \\
		P-GNN-F-2L + Hash &
		$0.978 \pm 0.003$ &
		$0.340 \pm 0.000$ &
		$0.988 \pm 0.004$ &
		$0.358 \pm 0.000$ &
		$0.769 \pm 0.031$ &
		$0.529 \pm 0.007$ &
		$0.734 \pm 0.033$ &
		$0.615 \pm 0.006$ \\
		P-GNN-F-2L + MSE &
		$0.982 \pm 0.004$ &
		$0.338 \pm 0.009$ &
		$0.991 \pm 0.002$ &
		$0.357 \pm 0.011$ &
		$0.835 \pm 0.016$ &
		$0.516 \pm 0.015$ &
		$0.772 \pm 0.016$ &
		$0.609 \pm 0.015$ \\
		P-GNN-F-2L + Both &
		$0.980 \pm 0.104$ &
		\boldmath{$0.341 \pm 0.072$} &
		$0.987 \pm 0.077$ &
		\boldmath{$0.359 \pm 0.071$} &
		$0.823 \pm 0.040$ &
		\boldmath{$0.532 \pm 0.056$} &
		$0.769 \pm 0.030$&
		\boldmath{ $0.616 \pm 0.006$}
		\\
				\hline
		P-GNN-E-2L &
		$0.965 \pm 0.002$ &
		$0.554 \pm 0.017$ &
		$0.981 \pm 0.000$ &
		$0.518 \pm 0.018$ &
		$0.792 \pm 0.031$ &
		$0.547 \pm 0.011$ &
		$0.735 \pm 0.047$ &
		$0.603 \pm 0.013$ \\
		P-GNN-E-2L + Hash &
		$0.973 \pm 0.006$ &
		\boldmath{$0.600 \pm 0.004$} &
		$0.981 \pm 0.001$ &
		$0.576 \pm 0.047$ &
		$0.770 \pm 0.030$ &
		$0.549 \pm 0.008$ &
		$0.784 \pm 0.020$ &
		\boldmath{$0.638 \pm 0.001$} \\
		P-GNN-E-2L + MSE &
		$0.985 \pm 0.110$ &
		$0.566 \pm 0.161$ &
		$0.990 \pm 0.092$ &
		$0.535 \pm 0.152$ &
		$0.765 \pm 0.025$ &
		$0.539 \pm 0.010$ &
		$0.772 \pm 0.014$ &
		$0.612 \pm 0.007$ \\
		P-GNN-E-2L + Both &
		$0.982 \pm 0.053$ &
		$0.588 \pm 0.013$ &
		$0.994 \pm 0.041$ &
		\boldmath{$0.607 \pm 0.010$} &
		$0.775 \pm 0.007$ &
		\boldmath{$0.550 \pm 0.005$} &
		$0.753 \pm 0.010$ &
		$0.614 \pm 0.000$
		\\
		\bottomrule
	\end{tabular}
}
	\end{center}
	\vspace{-0.6cm}
\end{table*}

\subsection{Datasets}
We conduct experiments on three real-world datasets -- Communities, Email, and PPI.
\begin{noindlist}
\vspace{-0.4cm}
\item \textbf{Communities}~\cite{watts1999networks} dataset is generated by first producing $n=20$ cliques of size $k=20$ each and then generating an edge to connect with an adjacent clique. To introduce some randomness, we connect cliques using an edge with a probability of $0.01$. Furthermore, we use node-clique membership as the natural label for the node pair classification task.
\vspace{-0.2cm}
\item \textbf{Email}~\cite{leskovec2007graph} is a collection of seven real-world email communication graphs with no node features. Each graph in this dataset has six communities and each node is labelled with the community it belongs to, which is used as the node pair classification label.
\vspace{-0.2cm}
\item \textbf{PPI}~\cite{zitnik2017predicting} network contains $1113$ nodes and each node is equipped with a $29$ dimensional feature vector. Each node represents a protein and an edge exists between two proteins if they interact with each other. Note that we only conduct the link prediction task for this dataset.
\vspace{-0.1cm}
\end{noindlist}

\vspace{-0.4cm}
\begin{table}[!htb]
	\caption{Results for PPI Dataset}
	\label{Exp:ppiExpResults}
	\begin{center}
	\resizebox{0.9\columnwidth}{!}{
		\begin{tabular}{ccc}
		\toprule
		\multicolumn{1}{c}{\multirow{2}{*}{Model Variants}} & \multicolumn{2}{c}{Link Prediction}                          \\ \cline{2-3} 
		\multicolumn{1}{c}{}                                & \multicolumn{1}{c}{AUC} & \multicolumn{1}{c}{Kendall's Tau} \\ \hline
		GCN               & $0.798\pm0.005$              & $0.413\pm0.010$             \\
		GCN + Hash        & $0.810\pm0.002$              & $0.441\pm0.007$             \\
		GCN + MSE         & $0.760\pm0.003$ & $0.422 \pm 0.005$ \\
		GCN + Both        & $0.821\pm0.001$ & \boldmath{$0.444 \pm 0.005$}\\
		\hline
		SAGE         & $0.809 \pm 0.004$ & $0.426 \pm 0.015$ \\
		SAGE + Hash  & $0.818 \pm 0.003$             & \boldmath{$0.464\pm0.012$}            \\
		SAGE + MSE   & $0.804 \pm 0.005$ & $0.421 \pm 0.012$ \\
		SAGE + Both  & $0.812 \pm 0.001$ & $0.461 \pm 0.002$ \\
		\hline
		GAT               & $0.798 \pm 0.007$ & $0.416 \pm 0.012$ \\
		GAT + Hash        & $0.818\pm0.015$             & $0.462\pm0.010$             \\
		GAT + MSE         & $0.789\pm0.004$                   &$0.413\pm0.012$                   \\
		GAT + Both        & $0.819 \pm 0.005$ & \boldmath{$0.465 \pm 0.012$} \\
		\hline
		GIN               & $0.755\pm0.014$                   &   $0.442\pm0.010$                \\
		GIN + Hash        & $0.788\pm0.024$          & $0.450\pm0.029$              \\
		GIN + MSE         & $0.751\pm0.012$                   &$0.432\pm0.045$                   \\
		GIN + Both        & $0.789\pm0.038$ & \boldmath{$0.460 \pm 0.053$}  \\
		\hline
		P-GNN-F-2L        &$0.812\pm0.007$                   &$0.375\pm0.001$                   \\
		P-GNN-F-2L + Hash &$0.823\pm0.004$                   &$0.388\pm0.005$                   \\
		P-GNN-F-2L + MSE  &$0.815\pm0.014$                   &$0.372\pm0.005$                   \\
		P-GNN-F-2L + Both &$0.819\pm0.007$             & \boldmath{$0.391\pm0.010$}           \\
		\hline
		P-GNN-E-2L        & $0.772 \pm 0.003$ & $0.435 \pm 0.004$ \\
		P-GNN-E-2L + Hash & $0.775\pm0.014$                  &$0.426\pm0.003$                   \\
		P-GNN-E-2L + MSE  & $0.792\pm0.015$                   &$0.446\pm0.005$                   \\
		P-GNN-E-2L + Both & $0.798\pm0.015$             & \boldmath{$0.454\pm0.005$}   
		\\
		\bottomrule       
	\end{tabular}}
\end{center}
\vspace{-0.7cm}
\end{table}

\subsection{Experiment Setup}
We consider the following baselines and their modifications to understand the impact of different modifications to the objective function used for graph embedding.
\begin{noindlist}
	\vspace{-0.3cm}
	\item Baseline Models: We consider GCN~\cite{Kipf:2016tc}, GraphSage (SAGE)~\cite{hamilton2017inductive}, Graph Attention Network (GAT)~\cite{gatpaper}, Graph Isomorphism Network (GIN)~\cite{xu2018how}, and P-GNN~\cite{DBLP:conf/icml/YouYL19} as our baseline models without modification of either node features or loss function.
	\vspace{-0.2cm}
	\item Baseline + Hash: For this category of variations, we only endow each node with hashing based features.
	\vspace{-0.2cm}
	\item Baseline + MSE: For this category of variations, we modify the objective function in Equation~\ref{eq:objective} with losses from the target task metric and distance distortion.
	\vspace{-0.2cm}
	\item Baseline + Both: This is the combined version that includes the modified objective function and the input representation extension.
\end{noindlist}
\vspace{-0.2cm}

\begin{figure}[htbp]
\centering
\includegraphics[width=0.43\textwidth]{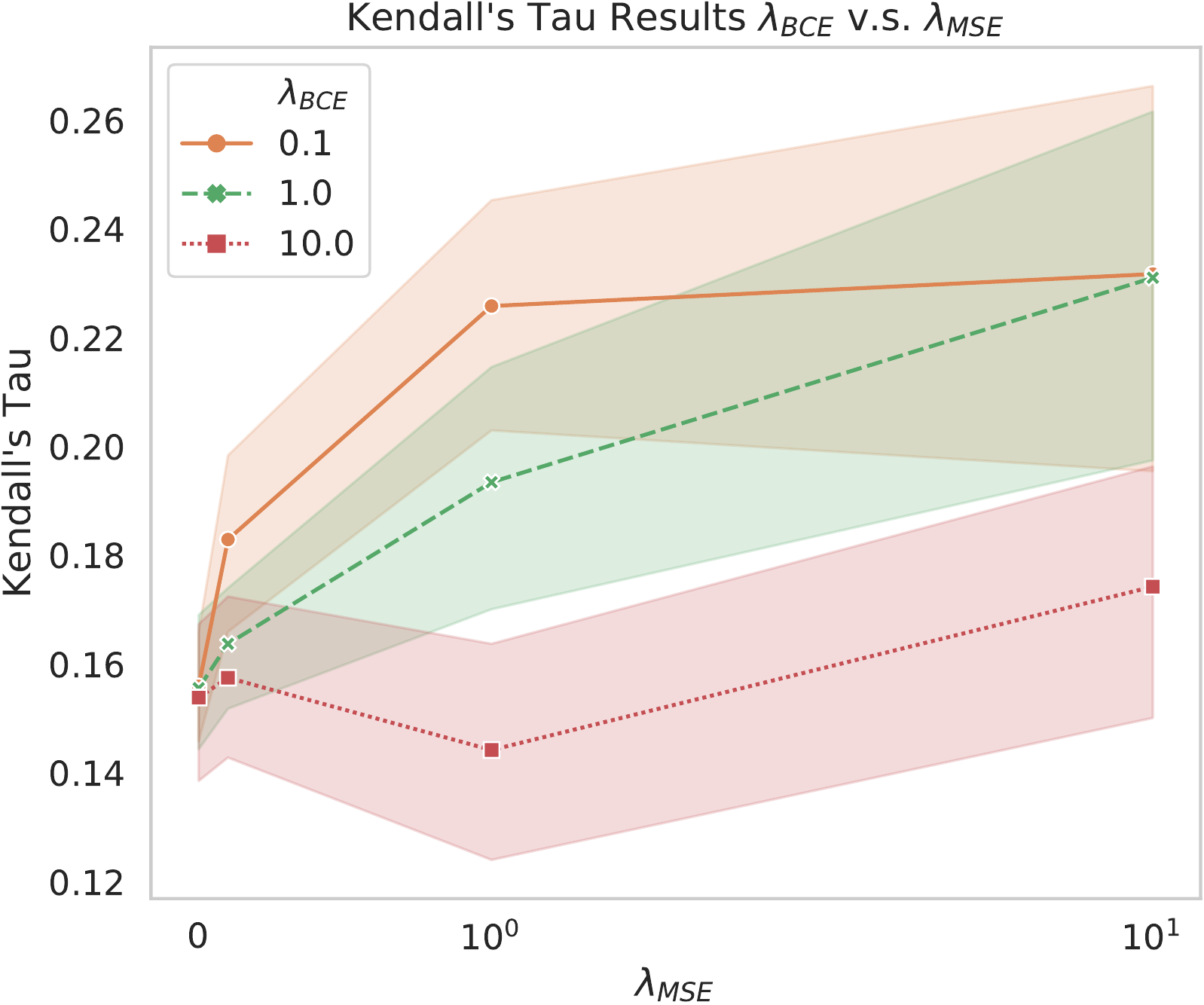}
\vspace{-0.2cm}
\caption{Impact of $\lambda_{\textrm{MSE}}$ on KT}
\label{fig:L1vsL2}
\vspace{-0.3cm}
\end{figure}

\begin{figure*}
\centering
\begin{subfigure}{.5\textwidth}
  \centering
  \includegraphics[scale=0.45]{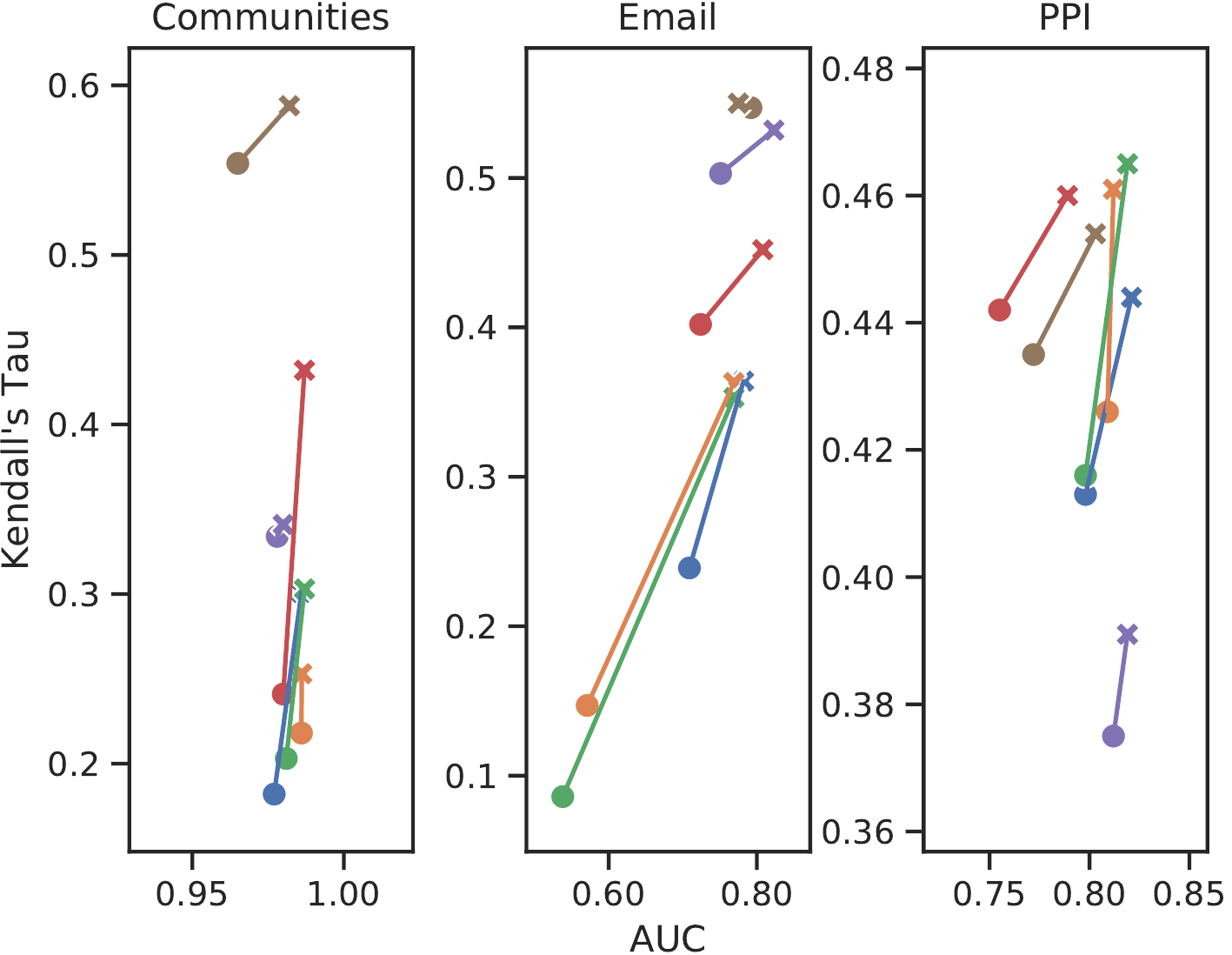}
\end{subfigure}%
\begin{subfigure}{.5\textwidth}
  \centering
  \includegraphics[scale=0.45]{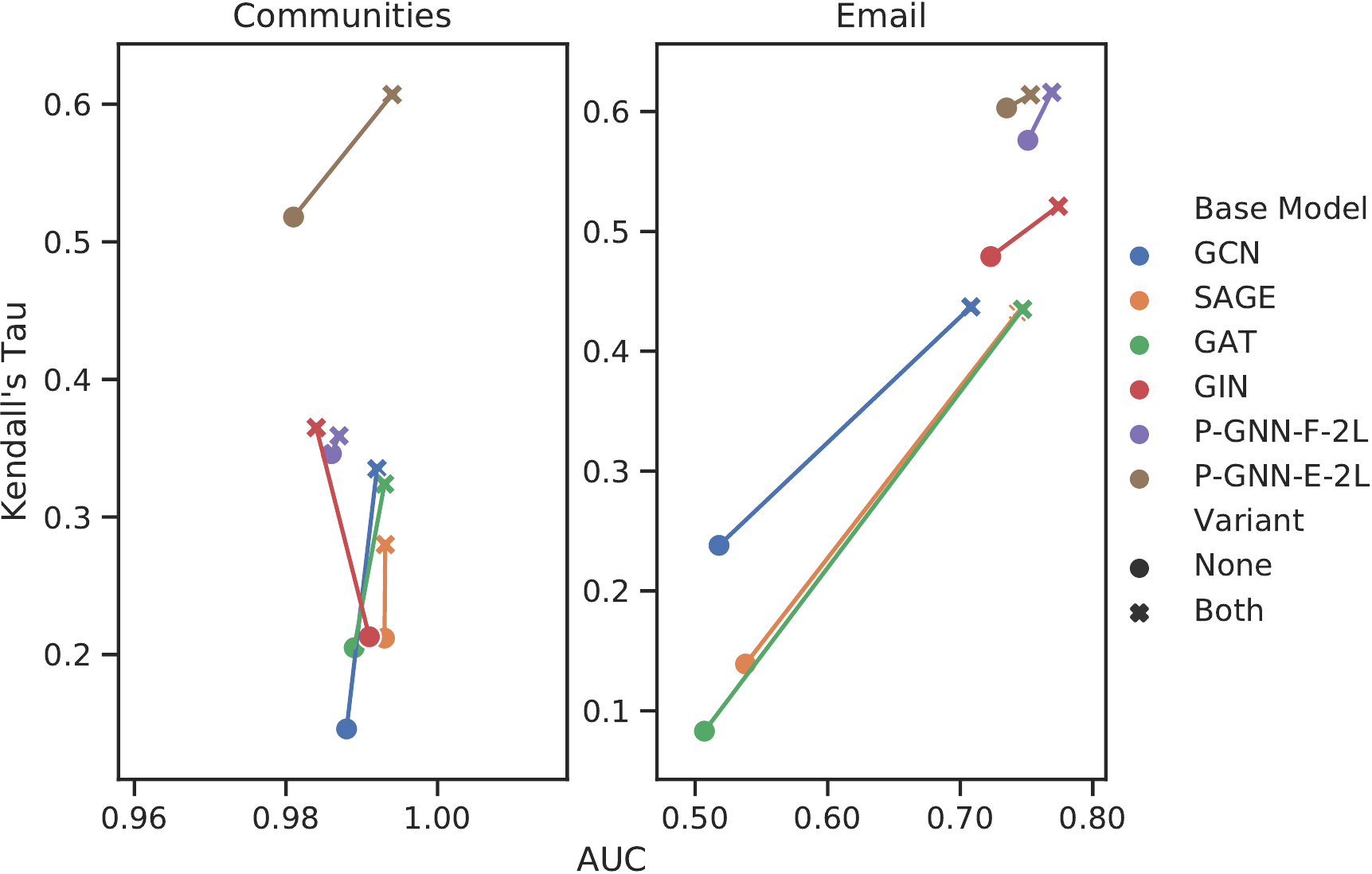}
\end{subfigure}
\vspace{-0.2cm}
\caption{AUC-ROC vs KT for Link Prediction (Left) and Pairwise Node Classification (Right)}
\label{fig:AUCvsKTNC}
\vspace{-0.3cm}
\end{figure*}

We evaluate the performance of all the models using the standard measure of classification performance, AUC-ROC, and a metric sensitive to distance, Kendall's Tau~\cite{kendallTau,kendallTauComputation}. Kendall's Tau increases as distortion decreases. We use a variant of the Kendall's Tau (KT) measure that adjusts for any ties. Given two lists of rankings, $A$ and $B$, this measure is calculated as 
\begin{equation}
\tau_b = \frac{(P - Q)} {\sqrt{(P + Q + T) \times (P + Q + U)}}
\end{equation}
where $P$ is the number of concordant pairs, $Q$ is the number of discordant pairs and $T$ ($U$) is the number of ties in $A$ ($B$) only. A tie that occurs for a pair in both $A$ and $B$ is not counted in either $T$ or $U$. We compare the rankings in the ascending order of graph distance and in the descending order of cosine similarity.

\subsection{Hyper-parameter Selection}
All of our experiments use \textsc{PyTorch}~\cite{pytorch2019} with models implemented, when available, in the \textsc{PyTorch-Geometric} package~\cite{pytorchGeometric}. For the GCN, SAGE, GIN, and GAT models, we use three hidden layers with $32$ dimensions. For P-GNN, we use two hidden layers (-2L) with either truncated 2-hop (-F) or exact (-E) shortest path distance. We perform a hyper-parameter search with three possible values of the learning rate $0.0001$, $0.001$, and $0.01$. The IGNN prescription introduces two new hyper-parameters -- the dimension $n$ of the hash vector and the strength $\lambda_{\text{MSE}}$ associated with the MSE objective. In practice, setting $n$ to the dimension of the observed node features works reasonably well. We also experiment with different values of $\lambda_{\text{BCE}}$ and $\lambda_{\text{MSE}}$ -- $0, 0.1, 1.0, 10.0$ -- to understand their impact on AUC-ROC and KT. Figure~\ref{fig:L1vsL2} demonstrates the effect of $\lambda_{\text{MSE}}$ on the KT measure for different values of $\lambda_{\text{BCE}}$ when GCN algorithm is applied on the Communities dataset. One can observe that reducing $\lambda_{\text{BCE}}$ and setting a slightly higher $\lambda_{\text{MSE}}$ produce the best KT results with minimal effect on AUC-ROC. We use the best hyper-parameter set from the Email dataset and use that in the PPI dataset, as sweeping through all possible configurations of the hyper-parameters is quite expensive for the PPI dataset. We report the test results using the hyper-parameter configurations that yield the best validation results.

\vspace{-0.1cm}
\subsection{Results}
We present the results for the Communities, Email, and PPI datasets in Tables~\ref{Exp:all} and~\ref{Exp:ppiExpResults}, respectively. Table~\ref{Exp:all} presents the AUC-ROC and Kendall's Tau measures for the Communities and Email datasets for both the link prediction and pairwise node classification tasks for the four variants of each of the six model architectures we experiment with. Table~\ref{Exp:ppiExpResults} contains the AUC-ROC and KT for PPI for the link prediction task for the same set of model variants.

It is obvious from these results that adopting the IGNN technique improves KT for virtually all the datasets, tasks, and architectures, by as much as $400\%$ in one case. In the Email dataset, we also see substantial improvement in the AUC-ROC, up to $43\%$. In the Communities dataset, the improvement is incremental, as most of the baseline models already have a very high AUC-ROC.

Figure~\ref{fig:AUCvsKTNC} shows how AUC-ROC and KT evolve between the baseline and the ``Both" variant of each model for the link prediction as well as the pairwide node classification task, allowing comparison of the improvement between different model architectures. In all cases, the IGNN models (``Both'' variants) have an increase in KT measurement. In more than 93\% (28 out of 30) of the cases, the AUC-ROC scores for the IGNN models are better than or as good as baseline models. 

When reviewing the results for the Communities and Email datasets, which use one-hot encoded input features, we see that GCN, SAGE, and GAT have similar KT for both prediction tasks for the baseline models. This result is not surprising, as these architectures focus primarily on incorporating node features and are known to be challenged in distinguishing certain structural differences. All three architectures register substantial improvements when they are empowered with the IGNN technique. The variants associated with the GIN and P-GNN have notably higher baseline performance, a result that is not surprising, as these architectures specifically aim to break structural isomorphisms and improve position awareness, respectively. Still, the improvement in KT is up to $70\%$ for GIN and $17\%$ for P-GNN variants.

In the PPI dataset where each node is attributed with a set of features, the situation is much different. The improvement in performance compared to the baselines is much lower than in the other datasets. The AUC-ROC improvement between baseline and the best IGNN variant is limited to $3\%$ only, though we see improvement up to $9\%$ in the KT measure. A notable exception is the P-GNN-F variant, which uses an approximate distance graph measure in the architecture (in contrast to the P-GNN-E variant, which uses the exact distance). This introduction of biased information likely results in lower performance. However, the fact that the architectures, such as GCN, SAGE, GAT, that focus primarily on node feature learning can achieve performance similar to the architectures explicitly taking advantage of structural or position information (GIN and P-GNN) strongly supports that the prescription presented in this paper enables these architectures to learn distance information.

\section{Conclusion}
We provide a prescription for training Isometric Graph Neural Networks using \textit{any} GNN architecture, allowing these algorithms to compute node representations that remain faithful to graph distance. This prescription introduces minimal complexity overhead, allowing different architectures to maintain their respective advantages. We show that IGNNs improve performance over baselines in a variety of tasks in both the standard metric of AUC-ROC used for link prediction and the distance sensitive metric Kendall's Tau.

\newpage
\clearpage
\small
\bibliography{reference}
\bibliographystyle{icml2020}

%
%
%

\end{document}